\documentclass[letterpaper, 10 pt, conference]{ieeeconf}

\IEEEoverridecommandlockouts                              
\def\NAT@parse{\typeout{IEEEtran error: Attempt to use fake Natbib command 
which is provided to fool Hyperref.}}

\usepackage{times}
\usepackage[ruled,vlined,linesnumbered]{algorithm2e}
\usepackage{multicol}
\usepackage[bookmarks=true]{hyperref}
\usepackage{graphicx} 
\usepackage{tabularx}
\usepackage{amssymb}
\usepackage{amsmath}
\usepackage{fancyhdr}
\usepackage{tikz}
\usepackage{bm}
\usepackage{multicol}
\usepackage{listings}
\usepackage{subfigure}
\usepackage{xcolor}
\usepackage{scalerel}
\usepackage[colorinlistoftodos]{todonotes}

\SetAlFnt{\small}
\SetAlCapFnt{\small}
\SetAlCapNameFnt{\small}

\newcommand{\matr}[1]{\mathbf{#1}} 

\newcommand{\vect}[1]{\mathbf{#1}} 
\newcommand{\vectg}[1]{\boldsymbol{#1}}

\newcommand{\bx}{\mathbf{x}}

\newcommand{\bu}{\mathbf{u}}

\newcommand{\bR}{\mathbf{R}}
\newcommand{\bT}{\mathbf{T}}

\newcommand{\bff}{\mathbf{f}}

\newcommand{\bex}{{\bf e}_{x}}

\newcommand{\bez}{{\bf e}_{z}}

\newcommand{\norm}[1]{{\|{#1}\|}}

\allowdisplaybreaks
\renewcommand{\baselinestretch}{0.98}
\begin{document}

\title{\LARGE \bf Dense Fixed-Wing Swarming using Receding-Horizon NMPC}
\author{Varun Madabushi$^{1}$, Yocheved Kopel$^{1}$, Adam Polevoy$^{1,2}$, and Joseph Moore$^{1,2}$
\thanks{$^{1}$Johns Hopkins University Applied Physics Lab \newline \hspace*{1.6em}
{\tt\small \{Varun.Madabishi,Yocheved.Kopel,\newline Adam.Polevoy,Joseph.Moore\}@jhuapl.edu} \newline \hspace*{0.8em} $^{2}$Johns Hopkins University Whiting School of Engineering \newline \hspace*{1.6em} {\tt\small }}}
\maketitle

\begin{abstract}
In this paper, we present an approach for controlling a team of agile fixed-wing aerial vehicles in close proximity to one another. Our approach relies on receding-horizon nonlinear model predictive control (NMPC) to plan maneuvers across an expanded flight envelope to enable inter-agent collision avoidance. To facilitate robust collision avoidance and characterize the likelihood of inter-agent collisions, we compute a statistical bound on the probability of the system leaving a tube around the planned nominal trajectory. Finally, we propose a metric for evaluating highly dynamic swarms and use this metric to evaluate our approach. We successfully demonstrated our approach through both simulation and hardware experiments, and to our knowledge, this the first time close-quarters swarming has been achieved with physical aerobatic fixed-wing vehicles. 
\end{abstract}
\IEEEpeerreviewmaketitle

\section{Introduction}

Swarms of uncrewed aerial vehicles (UAVs) are able to complete parallelizable tasks with greater speed and resilience than would be possible with a single robot. In particular, fixed-wing UAV swarms have the unique ability to leverage large lifting surfaces to achieve efficient flight and long range operations. However, to date, most UAV swarms have been comprised of multi-rotors, especially in cases where the UAVs must operate in close proximity to one another or other obstacles in the environment. By contrast, fixed-wing UAV swarms have often maintained large stand-off distances to avoid collisions. This is primarily due to the limited agility of fixed-wing UAVs. However, this limitation is not always due to the physical capabilities of the fixed-wing aircraft, but rather due to the inability of the underlying control algorithm to reason about the full flight envelope. In many instances, it would be advantageous to preserve both the long range capabilities of fixed-wing UAVs and the maneuverability of quadrotor UAVs.

In this paper, we present an approach for controlling multiple fixed-wing UAVs in close proximity to one another using receding-horizon nonlinear model predictive control (NMPC) so as to achieve both multi-rotor agility and fixed-wing range. Our approach relies on a direct transcription formulation of the trajectory optimization problem and is therefore able to encode both cost functions and constraints on the state variables. To prevent vehicle-vehicle collisions, we share trajectories among vehicles and formulate trajectory obstacle constraints. To ensure that vehicles stay close to their nominal trajectories, we employ a local time-varying feedback controller that is updated in real time, and characterize the performance of this feedback controller via statistical analysis to provide stochastic performance guarantees. 

We show that our approach is able to make use of a large portion of the flight envelope to enable close-quarters multi-vehicle operations. We demonstrate our approach in simulation and demonstrate multiple multi-vehicle patterns in hardware, with up to four fixed-wing UAVs flying in close proximity. We also compare our results to other swarms in the literature in terms of a ``swarm energy density'' metric. To our knowledge, this is the first time close-quarters swarming has been demonstrated in hardware with agile fixed-wing UAVs. 
\begin{figure}
    \centering
    \includegraphics[width=1\linewidth, trim={10mm 37mm 8mm 10mm}, clip]{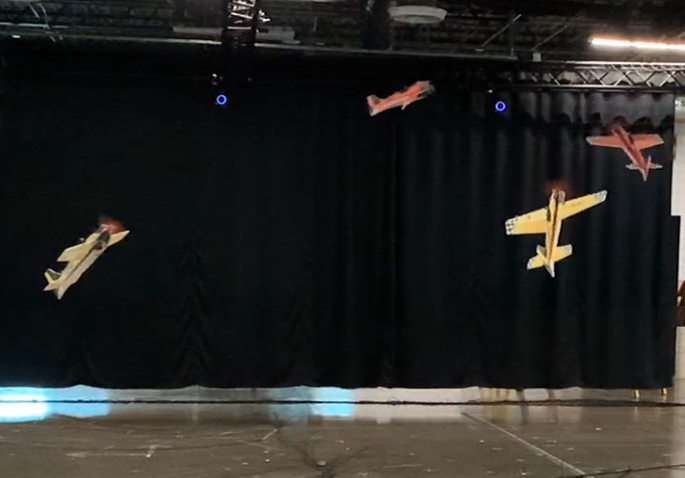}
    \caption{A still of four aerobatic fixed-wing UAVs executing alternating flight patterns in close proximity.}
    \label{fig:four-uav}
\end{figure}

\section{Related Work}
UAVs have been the subject of multi-agent control and swarming research for many years. Some of the earliest multi-vehicle control research began with coordinating teams of fixed-wing UAVs. In \cite{chalmers2004cooperating,bamberger2006flight}, the authors demonstrated cooperative fixed-wing UAV control via a potential field approach. \cite{bayraktar2004experimental} presents a multi fixed-wing UAV testbed and demonstrates formation control with two vehicles. \cite{how2004flight} demonstrates task assignment and receding-horizon trajectory generation. \cite{tisdale2008autonomous,tisdale2008multiple} use fixed-wing UAVs to execute autonomous search behaviors. \cite{sukkarieh2003anser} used fixed-wing UAVs for distributed data fusion. In \cite{chung2016live}, the authors demonstrate the launch of 50 fixed-wing UAVs. In all of this research, altitude separation has been extensively leveraged to achieve collision avoidance. 

Multi-rotors have also been used to conduct multi-vehicle swarming research. Because of their ability to achieve vertical take-off and landing, and their ease of control, swarms of multirotor UAVs have been more widespread. In \cite{hoffmann2004stanford,hoffmann2006distributed}, the authors develop a UAV testbed and conduct research on multi-UAV search behaviors. In comparison to fixed-wing UAVs, multi-rotor UAVs swarms have done more to reason about vehicle dynamics. This is due primarily to the algorithms taking advantage of a differentially flat representation of the vehicle dynamics that simplifies the trajectory optimization problem. In \cite{kushleyev2013towards} and \cite{preiss2017crazyswarm} the authors use differential flatness to control large swarms of micro quadcopters indoors. In \cite{Hamer2018GPUTraj} the authors leverage a GPU to compute trajectories for quadcopters in parallel, also making use of the differentially flat representation. In \cite{arul2019lswarm}, the authors demonstrate coupled task and motion planning for UAV swarms in complex environments. Other recent research has demonstrated multi-quadcopter flight in more unstructured environments \cite{zhou2022swarm,ahmad2021autonomous}. 

In recent years, nonlinear model predictive control has emerged as a promising method for controlling unmanned aerial vehicles to achieve achieve autonomous maneuvers. In addition to the prior research in quadrotor control (e.g., \cite{neunert2016fast,falanga2018pampc}), NMPC has demonstrated success in the control of fixed-wing UAVs \cite{basescu2020direct,mathisen2016non,mathisen2021precision,basescu2023precision} for agile maneuvering.  Authors have also begun to use NMPC to control quadcopter swarms. In \cite{soria2021predictive} the authors use NMPC to navigate a UAV team through an obstacle field. 

In this paper, we present an approach for controlling teams of fixed-wing UAVs in close proximity using nonlinear model predictive control. In particular, we use a direct trajectory optimization approach, which allows us to enforce path constraints on the the planned trajectories to de-conflict flight paths with other team members. 


\section{Approach}

\subsection{Dynamics Model}
We use the dynamics model first described in \cite{basescu2020direct} and revisit it here for convenience.
The state is given as $\vect{x} = \begin{bmatrix} \vect{r}, \vectg{\theta}, \vectg{\delta}, \delta_t, \vect{v}, \vectg{\omega}\end{bmatrix}^T$, where $\vect{r} =\begin{bmatrix} x_r,y_r,z_r\end{bmatrix}^T$ represents the position of the center of mass in the world frame $O_{x_r y_r z_r}$, $\vectg{\theta}= \begin{bmatrix} \phi, \theta, \psi\end{bmatrix}^T$ represents the set of $x-y-z$ Euler angles applied in a $ZYX$ intrinsic rotation sequence, and $\vectg{\delta} = \begin{bmatrix} \delta_{ar}, \delta_e, \delta_r \end{bmatrix}^T$ are control surface deflections for the right aileron, elevator, and rudder respectively. The left aileron deflection is treated as a dependent parameter, with $\delta_{al} = -\delta_{ar}$. $\delta_t$ is the magnitude of thrust from the propeller, $\vect{v} = \begin{bmatrix} v_x, v_y, v_z\end{bmatrix}^T$ is the velocity of the center of mass in the world frame, and $\vectg{\omega} = \begin{bmatrix}\omega_x, \omega_y, \omega_z\end{bmatrix}^T$ is the angular velocity of the body in the body-fixed frame $O_{xyz}$, and  $\vect{u}_{cs} = \begin{bmatrix} \omega_{ar}, \omega_{e}, \omega_{r}\end{bmatrix}^T$, where $\vect{u}_{cs}$ are the angular velocities of the control surfaces.
The equations of motion are written as
\begin{align}
\dot{\vect{r}} &= \bR_b^r \vect{v},\quad \dot{\vectg{\theta}} = {\bT}_{\omega}^{-1}{\boldsymbol \omega}\nonumber\\
\dot{\vectg{\delta}} &= \vect{u}_{cs}, \quad \dot{\delta}_t = a_t \delta_t + b_t u_t\nonumber\\
\dot{\vect{v}} &= \vect{f}/m - \vectg{\omega}\times \vect{v},\quad \dot{\vectg{\omega}} = \matr{J}^{-1}(\vect{m}-\vectg{\omega}\times\matr{J}\vectg{\omega})
\label{eq:eom}
\end{align}
Here $m$ is vehicle mass, $\matr{J}$ is the vehicle's inertia tensor with respect to the center of mass, $\vect{f}$ are the total forces applied to the vehicle in body-fixed coordinates, $\vect{m}$ are the moments applied about the vehicle's center of mass in body-fixed coordinates, and $\vect{u}_{cs}$ are the angular velocity inputs of the control surfaces.
$\bR_b^r$ denotes the rotation of the body-fixed frame with respect to the world frame, and ${\bT}_{\omega}$ is the transformation which maps the euler angle rates to an angular velocity in body-fixed frame.
The forces acting on the vehicle can be written in the body-fixed frame as
\begin{align}
\vect{f}= \sum_i[ \bR_{s_i}^b \vect{f}_{s_i} ] - mg{\bR_b^r}^T \bez + {\bR_{t}^b}\vect{f}_{t},
\end{align}
where $\vect{f}_{s_i}$ represents the force due to each aerodynamic surface in the surface frame $O_{x_s y_s z_s}$ and $\vect{f}_{t}$ represents the force due to the propeller and is given as $\vect{f}_{t}=\begin{bmatrix}\delta_{t} & 0 & 0\end{bmatrix}^T$. $g$ is the acceleration due to gravity. $\bR_{t}^b$ is the rotation matrix that defines the orientation of the thrust source with respect to the body-fixed frame. $\bR_{s_i}^b$ is the rotation matrix that defines the aerodynamic surface reference frame (in which the $z$ axis is the surface normal) with respect to the body-fixed frame. The aerodynamic surfaces used for this model are the wing, the horizontal and vertical fuselage, the horizontal and vertical tail, and the control surfaces. 

To model forces on the aerodynamic surfaces, we use
\begin{align}
\vect{f}_{s_i}=f_{n,s_i}\bez = \frac{1}{2}C_{n_i}\rho|\vect{v}_{s_i}|^2 S_i\bez,
\end{align}
where $\rho$ is the density of air, $S_i$ is the surface area, and $\vect{v}_{s_i}$ is the velocity of the $i^{th}$ aerodynamic surface in the surface frame given as
\begin{align}
&\vect{v}_{s_i}={\bR_{s_i}^b}^T(\vect{v}_b+\vectg{\omega}\times\vect{r}_{h_i} + \gamma_i \vect{v}_{bw})+...\\ \nonumber&({\bR_{s_i}^b}^T\vectg{\omega}+\vectg{\omega}_{s_i})\times\vect{r}_{s_i}.
\end{align}
Here $\vect{v}_b={\bR_b^r}^T \vect{v}$, $\vect{r}_{h_i}$ represents the displacement from the vehicle center of mass to a point on the aerodynamic surface that is stationary in the body-fixed frame. $\vect{r}_{s_i}$ represents the displacement from the hinge point to the surface center of pressure in the surface frame, and $\vectg{\omega}_{s_i}$ is the simple rotation rate of the aerodynamic surface in the aerodynamic surface frame. This is only non-zero for actuated surfaces. $\vect{v}_{bw}$ is the velocity due to the backwash of the propeller. It is approximated using actuator disk theory as
\begin{align}
\vect{v}_{bw}=\left[ \sqrt{\norm{\vect{v}_p}_2^2+\frac{2\delta_{t}}{\rho S_{disk}}} -\norm{\vect{v}_p}_2 \right] \bex,
\end{align}
where $S_{disk}$ is the area of the actuator disk and $\vect{v}_p$ is the freestream velocity at the propeller. $\gamma$ is an empirically determined backwash velocity coefficient. $C_{n_i}$ comes from the flat plate model in \cite{hoerner1985fluid} and is given as
\begin{align}
C_{n_i} = 2\sin\alpha_{s_i},
\alpha_{s_i} = \arctan{\frac{v_{s_i,z}}{v_{s_i,x}}},
\end{align}
where $\alpha_{s_i}$ is the angle of attack for the $i^{th}$ aerodynamic surface. $\vect{m}$ in the body-fixed frame can be given as
\begin{align}
\vect{m}= \sum_{i}(\vect{l}_{s_i} \times \bR_{s_i}^b\vect{f}_{s_i}),
\end{align}
where $\vect{l}_{s_i}=\vect{r}_{h_i}+\bR_{s_i}^b\vect{r}_{s_i}$ is the vector from the vehicle center of mass to the surface center of pressure in the body-fixed frame.
To model the thrust dynamics, we assume a first order linear model for the thrust where $a_t$ and $b_t$ are constants and $u_t$ is the normalized control signal for the motor, $u_t\in{[0,1]} $.


\subsection{Trajectory Generation}
To generate aircraft trajectories, we use a direct transcription approach. Because direct transcription includes control inputs and system states as optimization decision variables, our approach allows us to enforce collision constraints. Direct transcription also affords better numerical conditioning \cite{tedrake2009underactuated}. To transcribe the optimization problem, we apply an implicit integration constraint by using Simpson's integration rule \cite{pardo2016evaluating}. To improve computational performance, we exploit the sparse structure of the optimization problem via the Sparse Nonlinear Optimizer (SNOPT) \cite{Gill:2005:Snopt}. Our trajectory optimization problem is
\begin{equation}
\begin{aligned}
& \underset{\vect{x}_k, \vect{u}_k, h }{\text{min}}
& & \vect{\bar{x}}_N^T \vect{Q}_f \vect{\bar{x}}_N + \sum_k^{N-1}{ \vect{x}_k^T \vect{Q}  \vect{x}_k + \vect{u}_k^T \vect{R}  \vect{u}_k}\\
& \text{s.t.} 
& & \forall k \in [0,\ldots,N]\\
&
& & \vect{x}_{k} - \vect{x}_{k+1} + \frac{h}{6.0}(\vect{\dot{x}}_{k} + 4 \vect{\dot{x}}_{c,k} + \vect{\dot{x}}_{k+1})= 0\\
&
& & -\vectg{\delta}_f \le \vect{x}_{N}-\vect{x}_f \le \vectg{\delta}_f,~ -\vectg{\delta}_i \le \vect{x}_{0}-\vect{x}_i \le \vectg{\delta}_i\\
&
& & \vect{x}_{min}  \le \vect{x}_{k} \le \vect{x}_{max},~ \vect{u}_{min}  \le \vect{u}_{k} \le \vect{u}_{max}\\
&
&  & h_{min}  \le h \le h_{max},~ \\
&
&  & c(\bx_k, \mathcal{M})  \ge 0,
\end{aligned}
\end{equation}
where 
\begin{align}
\vect{\dot{x}}_{k} &= \vect{f}(t, \vect{x}_{k}, \vect{u}_{k}),~~
\vect{\dot{x}}_{k+1} = \vect{f}(t, \vect{x}_{k+1}, \vect{u}_{k+1})\nonumber\\
\vect{u}_{c,k} &= (\vect{u}_{k} + \vect{u}_{k+1}) / 2\nonumber\\
\vect{x}_{c,k} &= (\vect{x}_{k} + \vect{x}_{k+1}) / 2 + h (\vect{\dot{x}}_{k} - \vect{\dot{x}}_{k+1}) / 8\nonumber\\
\vect{\dot{x}}_{c,k} &= \vect{f}(t, \vect{x}_{c,k}, \vect{u}_{c,k}).
\end{align}
Here, $\vect{x}$ is the system state and $\vect{u}$ is the control input. $h$ is the time step where $h_{min}=0.001s$ and $h_{max}=0.2s$. $\vectg{\delta}_f$ and $\vectg{\delta}_i$ represent the bounds on the desired final and initial states ($\vect{x}_f$,$\vect{x}_i$), respectively. $N$ is the number of knot points and $\bar{\vect{x}}_N = \vect{x}_N-\vect{x}_f$. $\matr{Q}$, $\matr{R}$, and $\matr{Q}_f$ are cost weighting matrices. $c(\bx_k, \mathcal{M})$ is a collision constraint dependent on the state $\bx_k$ and an obstacle map $\mathcal{M}$ of the environment.
\subsection{Trajectory Tracking}
To control the aircraft to a trajectory between re-plans, we use discrete-time time-varying LQR (TVLQR). The policy is
\begin{align}
\bu_k &= \matr{K}_k(\bx_k-\bx^d_k)+\bu^d_k.
\end{align}
where the time-dependent feedback gain matrix $\matr{K}_k = (\matr{R}_c + \matr{B}_k^T\matr{S}_{k+1}\matr{B}_k)^{-1}(\matr{B}_k^T\matr{S}_{k+1}\matr{A}_k)$
and $\matr{S}(t)$ is computed by integrating
\begin{align}
\matr{S}_{k-1} = &\matr{A}_k^T\matr{S}_k\matr{A}_k-...\nonumber\\&\matr{A}_k^T\matr{S}_k(\matr{R}_c+\matr{B}_k^T\matr{P}_k\matr{B}_k)^{-1}\matr{B}_k^T\matr{P}_k\matr{A}_k+\matr{Q_c}
\end{align}
backwards in time from $\matr{S}_N=\matr{Q}_{fc}$. Here $\matr{A}_k=\frac{\partial \bff(\bx_k^d,\bu_k^d)}{\partial \bx_k}$ and $\matr{B}_k=\frac{\partial \bff(\bx_k^d,\bu_k^d)}{\partial \bx_k}$. $\bx_k^d$ and $\bu_k^d$ are the nominal trajectories. $\matr{Q}_c$ and $\matr{R}_c$ are the weighting matrices on state and action respectively.
\subsection{Trajectory Obstacle Constraints}
To achieve distributed multi-vehicle collision avoidance, we employ a strategy where each vehicle passes planned trajectories to its team members. Each aircraft stores the received trajectories in a trajectory obstacle map $\mathcal{M}_{\tau}$ and optimizes its own receding-horizon policy assuming that the plans of the other aircraft are fixed during the planning interval. The trajectory obstacle map is used to formulate a set of trajectory obstacle constraints. Let $\bx_k^j$ be the trajectories planned by vehicle $j$. The trajectory obstacle constraint can be written as 
\begin{align}
||\bx^i_k-\bx_l^j||-d \geq 0, \forall l\in[0,...,N], \forall j\in[1,...,M]
\end{align}
such that $i\neq j$. Here $M$ are the total number of aircraft, $r$ is the trajectory separation distance, and $k,l$ are the time indices of each airplane's trajectory in $\mathcal{M}_{\tau}$. 
\subsection{Nonlinear Model Predictive Control}
We combine our direct online trajectory optimization with trajectory tracking to create a receding-horizon NMPC algorithm. During every planning interval of length $T$ seconds, the controller executes the TVLQR feedback policy to follow the last planned trajectory while a new optimal trajectory is generated. To generate this new trajectory, at the beginning of the planning interval, the controller first simulates the closed-loop feedback policy forward for $T$ seconds and computes a new initial state estimate $\hat{\bx}_0$. The algorithm then computes the optimal trajectory and local feedback policy to be executed during the next planning interval from this initial condition. 
\subsection{Robust Selection of Collision Distance}
\label{RobustRho}

\begin{figure}
    \centering
    \includegraphics[width=.65\linewidth, trim={10mm 10mm 10mm 10mm}, clip]{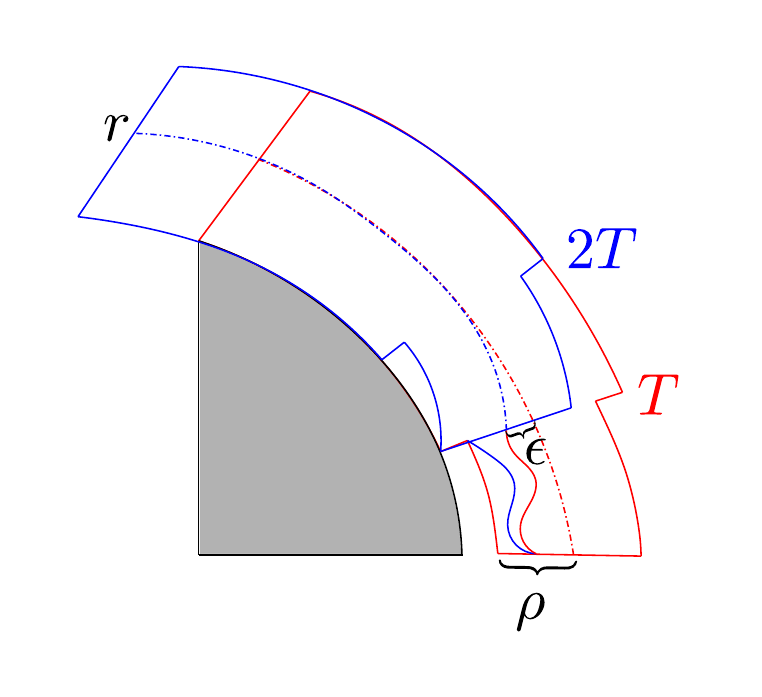}
    \caption{Figure shows how the trajectory tracking bound $\rho$ can combine with the re-planning deviation $\epsilon$ to increase the required collision radius.}
    \label{fig:samplebounds}
\end{figure}
Implementation of this planning algorithm for a particular robot swarm requires tuning of the parameter governing the allowable distance between agents. Our receding-horizon control approach generates trajectories for each agent which are separated from each other by a minimum distance $d$.
The local controller must then be able to keep the agent within a tube of radius $r=d/2-R$ of the planned trajectory, where $R$ is the physical robot radius. $r$ will be dependent both on the bound on trajectory tracking $\rho$ and the bound on the distance, $\epsilon$, between the current trajectory and the next planned trajectory, requiring a total collision radius of $r=\rho + \epsilon$ (see Figure \ref{fig:samplebounds}).


We propose a method which quantifies the tracking performance for an agent with stochastic dynamics based on recorded trajectory data. From this data, one can construct a confidence bound on the probability of collision, resulting in a safe
choice of $\rho$.

Since our LQR feedback controller is locally exponentially input-to-state stable, we assume that our system will be locally bounded in probability \cite{culbertson2023input} over a finite time interval according to $\mathbb{P}(||\bx(t)||\leq \rho) \geq (1-\upsilon)$ where $\upsilon$ is a small positive value. This property also holds for the positions $\bx_r(t)$, and we write $\mathbb{P}(||\bx_r(t)||\geq \rho) \leq \upsilon$. Violations of this condition can be represented as a random variable $X$, whose value is 1 if the trajectory leaves the tube and 0 if it stays within. Therefore, $\mathbb{E}[X]=P(||\bx_r(t)|| \geq \rho)$.
From repeated samples of $X$, we wish to find a bound $\mathbb{E}[X]$ with an appropriate confidence level. 
To do this, we leverage Hoeffding's Inequality, which relates the difference between the sample mean and true 
expectation of a bounded random variable \cite{alquier2023userfriendly} and is given as
\begin{align}
    \mathbb{E}[X] \leq \bar{X} + \sqrt{\frac{-\ln(\delta /2)}{2N}}.
\end{align}
Here, $\delta$ can be interpreted as confidence in the bound, which we set to $\delta=0.99$. $N$ is the number of samples, and $\bar{X}$ is the sample mean. Given $\delta$, $\rho$ can be selected such that the right hand side of the above inequality is sufficiently low, thus bounding the 
overall probability of constraint violation. To account for re-planning, we assume that $\epsilon$, the deviation between two successive planned trajectories will be bounded in probability over a finite time. This arises from the fact that the difference between the initial state of the next trajectory, $\hat{\bx}$, and the current trajectory will be bounded, since it arises from the forward simulation of a control policy for a deterministic system that is locally exponentially input-to-state stable. If our dynamics are Lipschitz continuous, there is a limit to how fast the trajectory can change over the next time interval. Therefore, we compute constraint violations for $||\bx_r(t)||\geq\rho$ for $t\in[0,T]$ and $||\bx_r(t)||\ge\rho+\epsilon$ for $t\in(T,2T]$, from the beginning of each re-planning interval, where $T$ is the re-planning interval. Given the uncertainty in the initial conditions and dynamics, we assume the samples in these intervals will be $i.i.d.$, but we note that the bounds may not hold under out-of-distribution data. 
\section{Simulation Results}
\label{sec:simresults}
To test the ability to bound the deviation from the nominal trajectory, we first ran a set of simulation experiments. In simulation, we collected trajectory data for a single 24-inch wingspan fixed-wing vehicle described in \ref{sec:setup} executing a box pattern. The box pattern is a reference path comprised of four straight lines, forming a 6.5m x 4m rectangle. We also collected data from three of the fixed-wing vehicles executing avoidance maneuvers while flying alternating box patterns. Each UAV travels along the box reference path at a constant velocity in alternating clockwise or counterclockwise directions, forcing the motion planner to resolve collisions regularly in each cycle.  The box patterns are offset vertically by $0.2m$ to avoid deadlocks due to trajectory optimization local minima. Our results showed that even a bound computed using a limited number of trials adequately bounded the long-term probability escaping the tube (see \ref{Sim Table}).
\begin{table}
\centering
\begin{tabular}{|c | c | c|}
\hline
\textbf{Maneuver} & \textbf{Bound}  & \textbf{Sample Mean}\\ 
\hline
\hline
1 plane box & 0.0517 (1031 trials) & 0.000263 (3799 trials) \\
\hline
3 plane box & 0.0314 (2997 trials) & 0.00320 (8117 trials) \\
\hline
\end{tabular}
\label{Sim Table}
\caption{Simulation analysis of performance bound.}
\end{table}
\begin{figure}
    \centering
        \includegraphics[width=1\linewidth, trim={0mm 0mm 0mm 0mm}, clip]{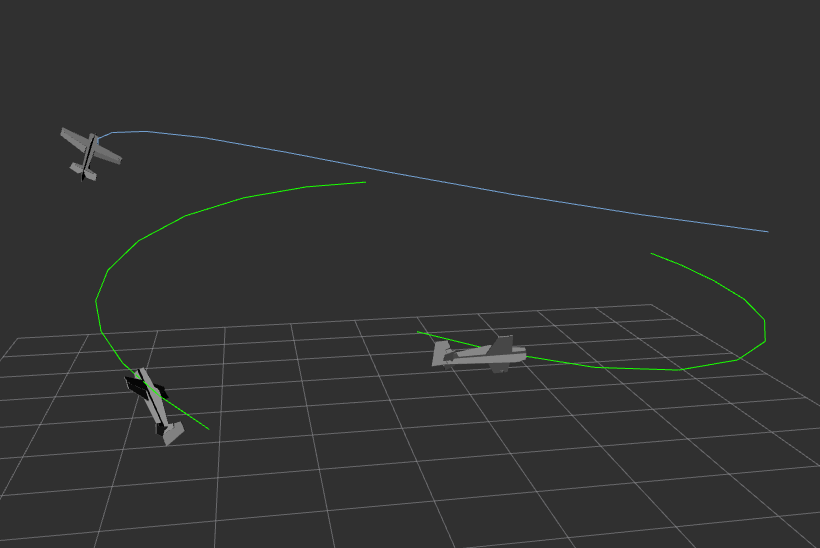}
    \caption{A still from a three-plane alternating box pattern simulation experiment.}
    \label{fig:uav-sim}
\end{figure}
\section{Experimental Results}
\begin{figure}
  \centering
  \includegraphics[width=1\linewidth]{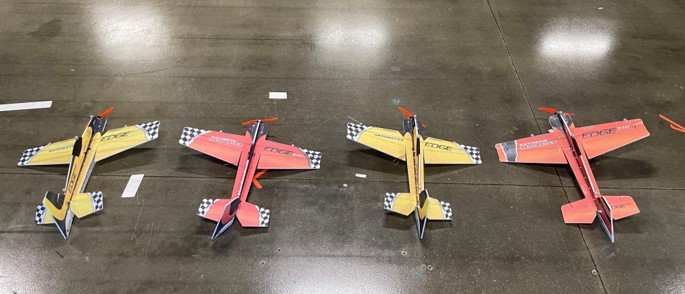}
\caption[Experimental Aircraft]{24" wingspan Edge 540 EPP model.}
  \label{Fig:edge540}
  \end{figure}

Given the fidelity limitations of the dynamics model used for control, real-world experiments were imperative in validating the efficacy of the controller design. We first conducted a series of experiments using a small, 24-inch wingspan foam aircraft (Figure \ref{Fig:edge540}) operating in an indoor motion-capture space. For these experiments, all state estimation and control computation was performed offboard the vehicle in real-time.
\subsection{Experimental Set-up}
\label{sec:setup}
The plane used for the experiments was a 120g, 24-inch wingspan aerobatic Edge 540 model produced by Twisted Hobbys (Figure \ref{Fig:edge540}). It has an independently controllable rudder and elevator, as well as a pair of kinematically linked ailerons. A 13g 2300Kv 2202 Crack Series out-runner motor was paired with a 7x3.5 GWS propeller for the thruster. In addition to the stock components for the aircraft, five retroreflective markers were attached at various locations on the airframe. 
Maximum throws were measured for each control surface along with their corresponding PWM inputs in order to generate a linear mapping between desired control surface angle and PWM value. To generate the desired deflections, we integrate angular velocity commands from the controller. 

The experiments were performed in an 8m x 8m capture area, with OptiTrack PrimeX 22 cameras distributed along the periphery of the space. The OptiTrack system was configured to report position and quaternion data at 100Hz to the computer running the planning and control algorithms, which was a Dell Precision 5530 laptop with an Intel i7-8850H CPU. The planning machine used the Robot Operating System (ROS) package \texttt{vrpn\_client\_ros} to generate a ROS topic with timestamped pose data, which was then differentiated over a single timestep to obtain linear and angular velocities. In addition, the quaternion data was transformed into yaw/pitch/roll euler angles.  

In order to deliver desired control inputs to the plane, a serial interface was developed to communicate with an 8MHz Arduino Pro Mini microcontroller. The planning computer sends the desired PPM signal to the microcontroller, which relays this signal to a Spectrum DX6i transmitter module through the trainer port.  The transmitter then broadcasts the signal to the RC receiver on the plane.  
For all experiments, the planes were launched by hand.
The control system is configured to automatically activate once it has crossed a position threshold after being thrown by the operator.

\subsection{Experiments}
We tested our algorithm on three sets of maneuvers: a 3-plane box pattern, a 3-plane search pattern, and a 4-plane box pattern, The box patterns follow the configuration described in \ref{sec:simresults}. The search pattern is a sequence of randomly sampled points from a 5.5m x 5.5m square grid. Traveling between subsequent points requires the UAV to make tight turns, and the UAVs paths may intersect at any angle, requiring the motion planner to resolve both head-on and cross collisions.

\begin{figure}
    \centering
    \includegraphics[width=\linewidth]{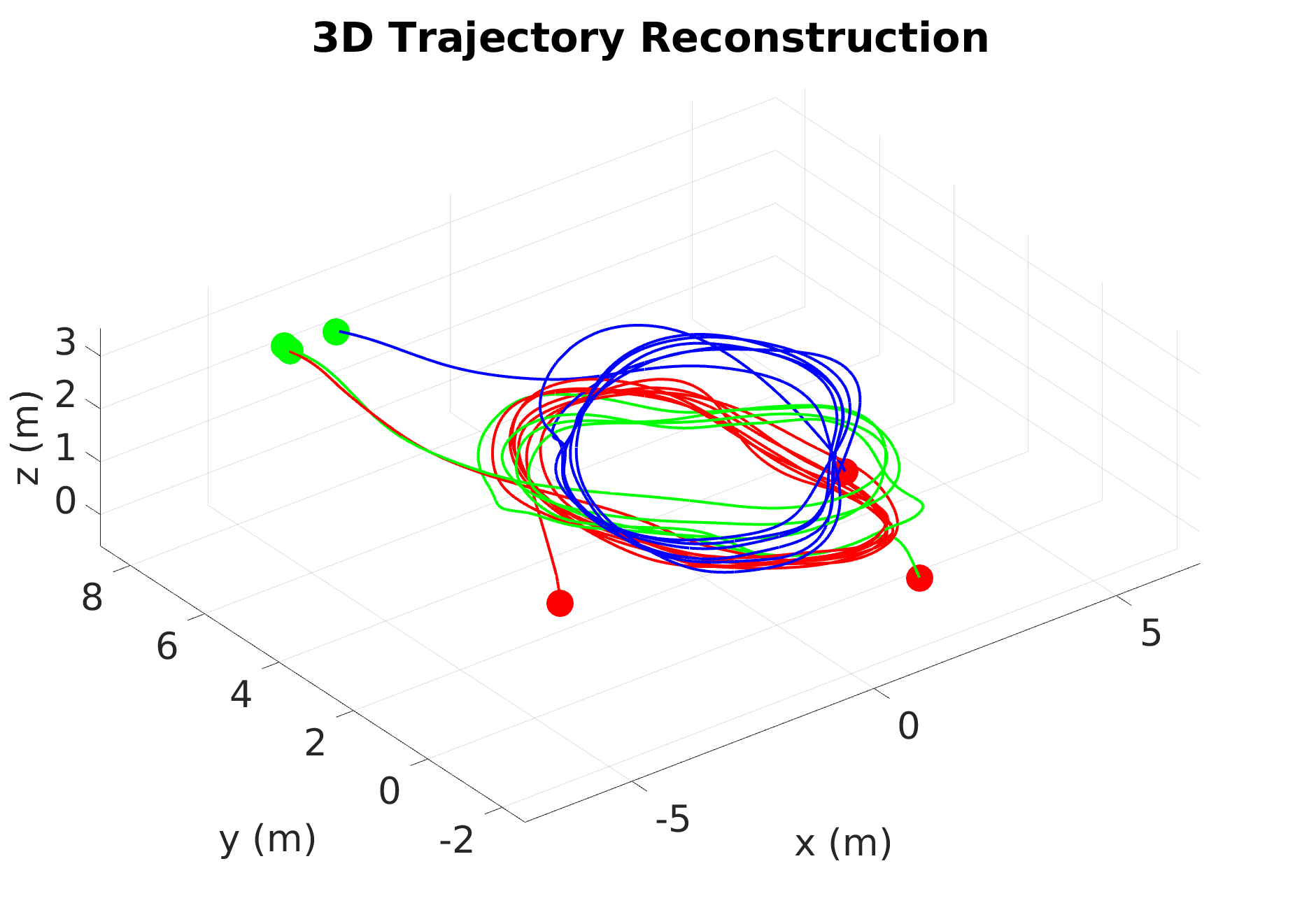}
    \caption{Trajectories of 3 planes flying in a box pattern}
    \label{fig:3DTraj3Plane}
\end{figure}

\begin{figure}
    \centering
    \includegraphics[width=\linewidth]{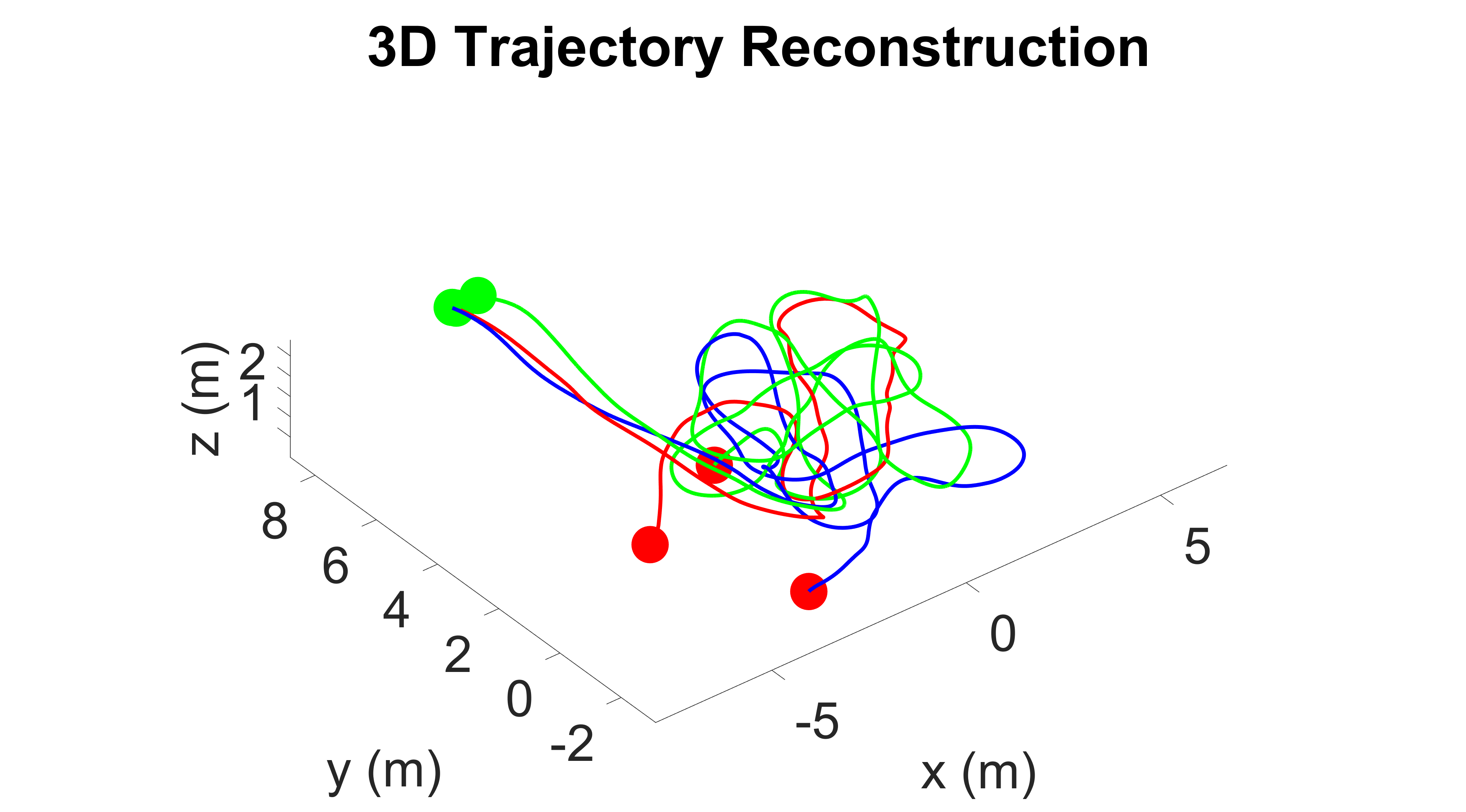}
    \caption{Trajectories of 3 planes flying in a search pattern}
    \label{fig:3DTraj3PlaneSearch}
\end{figure}

The trajectories of the three aircraft in a 3-plane box maneuver is shown in figure \ref{fig:3DTraj3Plane}.
Similarly, the trajectories of three aircraft in a search maneuver is shown in figure \ref{fig:3DTraj3PlaneSearch}. For both types of maneuvers, the agents were given a constant velocity command of 4m/s. This velocity is factored into the cost calculation used by the receding-horizon planner.
\begin{figure}
    \centering
    \includegraphics[width=1\linewidth]{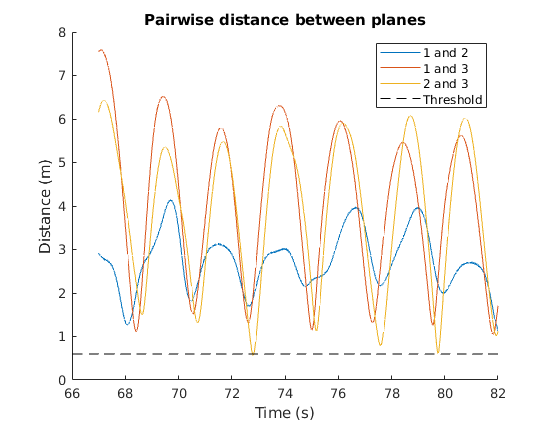}
    \caption{Distance between each pair of UAVs during a 3-plane box maneuver. The dashed line represents the distance where two UAVs are likely to crash into each other.}
    \label{fig:PairwiseDistSept6_2}
\end{figure}
We analyze the data gathered from flight trials to validate our choice of $\rho=0.35m, \epsilon=0.15m$, used by the controller to maintain safe separation between planes.
Our data set consists of 28 flights, for a total of 3,284 individual trajectories.
We generate a bound using the methodology specified in section \ref{RobustRho}, from a random selection of 27 flights.
The remaining flight is used for cross-validation of the generated bound.
We find that across our dataset, the plane stayed within $\rho$ of its commanded trajectory and replanned within $\rho + \epsilon$ of its commanded trajectory for 100\% of the trials.

\section{Analysis}
\subsection{Collision Risk}
To determine if UAVs were at risk of colliding, we can measure the distances between each pair of UAVs in the swarm.
This data, plotted for a  3 Plane Box maneuver, is shown in Figure \ref{fig:PairwiseDistSept6_2}.
The planes are able to fly as close as 0.55m from each other, which is the wingspan of a single plane. 
Table \ref{DistancesBetweenAgentsTable} shows the average and minimum distance between the closest pair of planes for each 
scenario.

\begin{figure}
    \centering
    \includegraphics[width=1\linewidth]{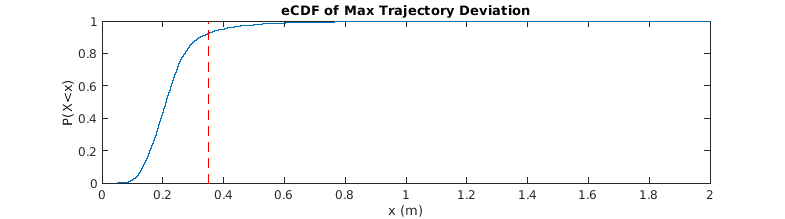}
    \caption{eCDF of trajectory positional error for all trajectories in the dataset. The red line represents a tracking error bound of $\rho=0.35m$.}
    \label{fig:TrajDevECDF}
\end{figure}

\begin{figure}
    \centering
    \includegraphics[width=1\linewidth]{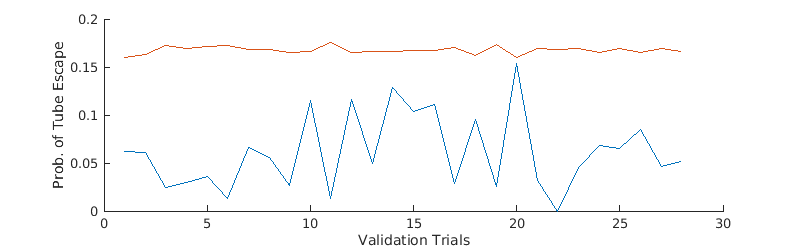}
    \caption{Cross-validation trials checking the trajectory deviation bound $\rho+\epsilon$. The sample means for the validation trials (blue) do not exceed the bound on the probability of leaving the tube (red).}
    \label{fig:validation}
\end{figure}

We analyze the data gathered from flight trials to validate our choice of $\rho$. 
Our data set consists of 28 flights, for a total of 3,284 individual trajectories.
We record each planned trajectory and the corresponding trajectory tracked by the aircraft, and measure the norm of the positional error, $||p-p_{des}||$.
The empirical cumulative distribution function of the error is shown in Figure \ref{fig:TrajDevECDF}.
93\% of trajectory tracking errors are less than our selected value of $\rho=0.35m$. 
Note that this is a conservative bound, since an agent could violate the bound and still maintain the minimum required separation between airplanes. We generate a bound using the methodology specified in section \ref{RobustRho}, from a random selection of 27 flights.
The remaining flight is used for cross-validation of the generated bound. We find that across our dataset, the average number of deviated trajectories was within the computed bound for all trials (see Figure \ref{fig:validation}).

In total, during our experimental trials, we flew three or more aircraft in close proximity for 113s, during which we experienced 86 pairwise crossings and 3 collisions, only one of which was fatal. We experienced 5 altitude violations, which was often the basis for the termination of the trial. These altitude violations were of due to low battery voltage or exceeding pitch constraints. 

\subsection{Swarm Energy Density}
In order to characterize the performance of swarming robotic systems in complex maneuvers, we introduce a metric which we call ``Swarm Energy Density,'' or SED.
This metric can be used to score a swarming robotic system in terms of its speed, number of agents, and distance between agents. The metric is given as $\frac{\sum_i m_iv_i^{2}}{2V}$ where $m_i$ is the mass of the $i^{th}$ agent, $v_i$ is the velocity of the $i^{th}$ agent, and $V$ is the volume of the 
smallest convex hull enveloping the swarm.

We compute SED by approximating a representative fixed-wing scenario similar to \cite{chung2016live}, a quadcopter scenario similar to \cite{zhou2022swarm,preiss2017crazyswarm,kushleyev2013towards}, and our own scenario (see \ref{ParamTable}). We also compute a more accurate time-average (TA) of the instantaneous SED from our trial data. The SED results are summarized in in Table \ref{EnergyDensityTable}.
\begin{table}
\centering
\begin{tabular}{| c | c | c | c | c |}
\hline
\textbf{Scenario} & \textbf{Volume} & \textbf{Speed} & \textbf{Number} &\textbf{Mass}\\ 
\hline
\hline
Fixed-Wing & 375mx375mx375m & 18m/s & 25 & 2.5kg \\
\hline
Quadcopter & 6mx6mx1m & 2m/s & 10& 0.3kg\\
\hline
Ours & 7mx7mx1.5m & 4m/s & 4& 0.12kg\\
\hline
\end{tabular}
\label{ParamTable}
\caption{Scenario Parameters}
\end{table}
\begin{table}
\centering
\begin{tabular}{|c c|}
\hline
\textbf{Swarm} & \textbf{Avg Energy Density} ($\bm{J/m^{3}}$) \\ 
\hline
\hline
Fixed-Wing & 0.000192 \\
\hline
Quadcopter & 0.167 \\
\hline
Ours (Approx) & 0.0522 \\
\hline
Ours TA (3 plane search) & 1.044 \\
\hline
Ours TA (3 plane box) & \textbf{1.160} \\
\hline
Ours TA (4 plane box) & 0.310 \\
\hline
\end{tabular}
\label{EnergyDensityTable}
\caption{Energy Density computed for various swarming systems}
\end{table}
The SED of our system shows a 2-3 order of magnitude increase over other fixed-wing swarms and is on-par with quadcopter swarms. The 3-plane box trial has a higher time-average SED since the three planes happen to often all overlap in time.
\begin{table}
\centering
\begin{tabular}{|c | c | c|}
\hline
\textbf{Scenario} & \textbf{Avg Density} ($\bm{Kg/m^{3}}$)  & \textbf{Max Density} ($\bm{Kg/m^{3}}$)\\ 
\hline
\hline
3 plane search & 0.1063 & 0.3874 \\
\hline
3 plane box & 0.0955 & 0.2950 \\
\hline
4 plane box & 0.0453 & 0.0783 \\
\hline
\end{tabular}
\label{Density Table}
\caption{Density of various maneuvers}
\end{table}

\begin{table}
\centering
\begin{tabular}{| c | c | c | c |}
\hline
\textbf{Scenario} & \textbf{Avg. Min. Distance} & \textbf{Min. Distance} &\textbf{Avg. AoA} \\ 
\hline
\hline
3 plane search & 2.2705m & 0.2984m & -0.4934 rad\\
\hline
3 plane box &  2.4498m & 0.5967m& -0.3457 rad\\
\hline
4 plane box & 1.8259m & 0.5567m& -0.5063 rad\\
\hline
\end{tabular}
\label{DistancesBetweenAgentsTable}
\caption{Minimum Distance between agents for various maneuvers}
\end{table}


\section{Discussion}
We described an approach for coordination and collision avoidance with a swarm of fixed-wing UAVs. Our approach leverages nonlinear model predictive control, which plans across a large flight envelope and uses high angle-of-attack maneuvers to avoid collisions. We successfully demonstrated tight maneuvers with up to 4 UAVs flying within 3 wingspans of each other at an average speed of 4m/s.


\bibliographystyle{IEEEtran}
\clearpage
\bibliography{references}

\end{document}